\documentclass[12pt, final]{l4dc2025}
\usepackage{xz_style}

\usepackage{graphicx}
\usepackage{wrapfig}
\usepackage{booktabs} 

\usepackage{url}
\usepackage{algorithm}
\usepackage{makecell}
\usepackage{algorithmic}
\usepackage{amssymb,amsmath}
\usepackage{subcaption}
\usepackage{comment}
\usepackage{thmtools, thm-restate}
\usepackage{multirow}

\usepackage{xcolor} 
\definecolor{ocre}{RGB}{243,102,25}

\usepackage{avant}
\usepackage{mathptmx} 
\usepackage{microtype}
\usepackage[utf8]{inputenc}
\usepackage{tcolorbox}

\begin{document}
\title{Learning Ensembles of Vision-based Safety Control Filters
}
\author{%
 \Name{Ihab Tabbara} \Email{i.k.tabbara@wustl.edu}\\
 \addr Washington University in St. Louis
 \AND
 \Name{Hussein Sibai} \Email{sibai@wustl.edu}\\
 \addr Washington University in St. Louis
}
\maketitle

\begin{abstract}
Safety filters in control systems correct nominal controls that violate safety constraints.
Designing such filters as functions of visual observations in uncertain and complex environments is challenging. Several deep learning-based approaches to tackle this challenge have been proposed recently.  
However, formally verifying that the learned filters satisfy critical properties that enable them to guarantee the safety of the system is currently beyond reach.   
Instead, in this work, motivated by the success of ensemble methods in reinforcement learning, we empirically investigate the efficacy of ensembles in enhancing the accuracy and the out-of-distribution generalization of such filters, as a step towards more reliable ones. 
We experiment with diverse  
pre-trained vision representation models as filter backbones, training approaches, and output aggregation techniques. We compare the performance of ensembles with different configurations against each other, their individual member models, and large single-model baselines in distinguishing between safe and unsafe states and controls in the DeepAccident dataset. Our results show that diverse ensembles have better state and control classification accuracies compared to individual models.
\end{abstract}
\begin{keywords}
Safety filters; Ensembles; Control barrier functions; Pre-trained vision models
\end{keywords}

\section{Introduction}

Ensuring safety of control systems is a fundamental challenge in various application domains, including autonomous driving~(\cite{autonomous_driving_a_crash_explained_in_detail_2019}), aerospace~(\cite{spacecraft_docking_cbf_panagou_2022}), and robotic surgery~(\cite{robotic_surgery_survey_2019}).  
It entails verifying that the trajectories of a system remain in a region of the state space that  the user considers safe, or synthesizing controllers that drive the system to remain there. One of the prominent solutions is to design barrier certificates that guarantee the safety of the system. These certificates can guide the selection of controls or modify nominal ones to maintain safety, effectively serving as safety control filters (\cite{ames2016control}). Unfortunately, synthesizing such certificates is generally NP-hard~(\cite{clark2021verification}) and accordingly, existing algorithms do not scale beyond few dimensions. Moreover, these algorithms require white-box settings where the dynamics of the system and its environment are known. For many modern systems, e.g., vision-based autonomous navigation, such conditions are not satisfied.   

Recently, deep learning-based methods have been proposed for  designing certificates and controllers, offering an easier and more scalable approach for their design~(\cite{learned_certificates_survey_chuchu_2023,abdi2023safe,tong2023enforcing,xiao2022differentiable,ICRA_2025_pre-trained_vision_cbfs}). However, the learned neural certificates are not formal ones. They do not necessarily satisfy the required conditions at every state and control for them to guarantee safety. Algorithms for formally verifying them suffer from similar curse-of-dimensionality limitations as the traditional methods for their design, despite significant progress in neural networks (NN) verification over the past few years~(e.g., \cite{katz2022verification,neural_lyapunov_control_for_discrete_systems_andrew_yiannis_eugene_Neurips_2023,NN_verification_book_aws_2021,alpha_beta_crown_2024}). 
~
Similarly, verifying the safety of systems with NN-based controllers, particularly vision-based ones, suffer from the same scalability challenges and is usually constrained to predefined simple environments or particular images and their local neighborhoods~(e.g., \cite{santa2022nnlander,hsieh2022verifying,cai2024scalable}). 

In this work, we take an alternative approach to formal verification and investigate using ensemble learning to improve the reliability and accuracy of vision-based safety filters.
Ensemble learning has been used for uncertainty quantification (\cite{rahaman2021uncertainty}), accuracy improvement, and out-of-distribution generalization~(\cite{sagi2018ensemble}), in various machine learning (ML) tasks, but has not been used for safety control filters design before, up to our knowledge.  We build the member models of our ensembles using the approach we presented in \cite{ICRA_2025_pre-trained_vision_cbfs}, which uses pre-trained vision representation models (PVRs), such as CLIP~(\cite{clip}) and VC1~(\cite{vc1}),  as perception backbones for the safety filters, significantly decreasing the sample complexity of learning the filters while improving generalization.
We focus on the  vision-based collision avoidance task in autonomous driving as the application domain. We use the DeepAccident dataset (\cite{wang2024deepaccident}), generated using CARLA (\cite{dosovitskiy2017carla}), to train and evaluate the  filters. 
We experiment with diverse ensembles which have member models with different PVR backbones, training methods, and model aggregation techniques.
We analyze the trade-off between performance and complexity. 
Our results show the advantages of using diverse ensembles  
instead of individual models, 
showing their potential as more reliable safety control  filters.

\section{Related Work}

\paragraph{Ensembles in reinforcement learning (RL) and control}
Ensembles have been used to represent and tackle  uncertainty in risk-sensitive RL~(\cite{eriksson2022sentinel,hoel2023ensemble}), 
for learning from unstable estimations of value functions~(\cite{fausser2015neural,anschel2017averaged}), for learning value functions more efficiently~(\cite{chen2021randomized}), to facilitate optimism for efficient exploration in model-based online deep RL~(\cite{pacchiano2021towards}), to enable pessimism in offline RL (\cite{ghasemipour2022so}), to approximate reward functions in inverse RL (\cite{lin2020ensemble}), for robust  dynamic motion prediction (\cite{mortlock2024castnet}), and for anomaly detection (\cite{ji2024expert}). Moreover, it has been shown that carefully designed reward functions  define Q functions that are equivalent to control barrier functions (\cite{value_functions_are_control_barrier_functions}), the control version of barrier certificates. This implies that the demonstrated benefits of ensembles in RL can also be potentially obtained in learning safety filters. The results of this paper can be seen as a supporting evidence.

\noindent \textbf{Learning safety filters}
Recent approaches to designing safety filters use deep learning to overcome the scalability challenges inherent in methods based on sum-of-squares optimization and reachability analysis~(\cite{dawson2022safe,dawson2022learning}) and  to account for unknown dynamics~(\cite{qin2022SABLAS,lavanakul2024safety,castaneda2023distribution}) and high-dimensional observations such as images and point clouds~(\cite{tong2023enforcing,abdi2023safe,barriernet_2023}). The resulting filters are not guaranteed to satisfy the conditions for them to be valid certificates, unless under  restrictive assumptions of known Lispchitz constants of the NNs and corresponding grid-like training datasets that cover the whole domain~(\cite{formally_verified_CBF_majid_2023,learning_formally_verified_CBF_stochastic_systems_andrew_2024}). Several works have used NN  verification techniques to guarantee these conditions as well as generating counter examples that can be used for retraining~(\cite{neural_lyapunov_control_for_discrete_systems_andrew_yiannis_eugene_Neurips_2023,verification_of_CBF_Changliu_CoRL_2024}). However, these techniques are not yet scalable enough to verify high-dimensional, observation-based filters across all possible scenarios,
e.g., those that can be observed in autonomous driving settings, and are instead constrained to local structured environments (\cite{abdi2023safe}). Instead, we aim to improve the empirical performance and tackle the epistemic uncertainty of vision-based safety filters through ensemble learning.

\section{Preliminaries}\label{sec:preliminaries}
In this section, we recall the definition of control barrier functions (CBF) and the guarantees they provide. We generally assume that the dynamics of the control system under consideration is control-affine. This assumption, while not required for the CBF definition, is usually added to obtain state-dependent linear constraints that separate safe and unsafe controls, which simplify the safety filtering optimization problem to a quadratic program that can be solved efficiently in real-time.

\begin{definition}[Control-affine control systems]
    A continuous-time nonlinear control-affine system can be described using the following ordinary differential equation (ODE):
    \begin{equation}\label{eq:system}
        \dot{x} = f(x) + g(x)u,
    \end{equation}
    where \( x \in \mathcal{X} \subset \mathbb{R}^n \) is the state variable, and \( u \in \mathcal{U} \subset \mathbb{R}^m \) is the control one. We assume that \( f: \mathcal{X} \to \mathbb{R}^n \) and \( g: \mathcal{X} \to \mathbb{R}^{n \times m} \) are locally Lipschitz continuous.
\end{definition}


\begin{definition}[Control barrier functions (\citet{cbf_overview_2018})]
    A continuously differentiable function \( B: \mathcal{X} \to \mathbb{R} \) is called a \emph{control barrier function
    } for system \eqref{eq:system}  if 
    \begin{align}
        \exists u \in \mathcal{U} \text{ such that } \dot{B}(x,u) + \gamma (B(x)) \geq 0 \quad \forall x \in \mathcal{D} \subseteq \mathcal{X}, \label{eq:cbf-def-cond-ascent}
    \end{align}
    where 
    the super-level set $B_{\geq 0}:=\{x\ |\ B(x) \geq 0\}$ of $B$ is a subset of $\mathcal{D}$,  and \( \gamma: (-b,a) \to \mathbb{R} \), for some $a,b > 0$, is a locally Lispchitz extended class \( \mathcal{K}_{\infty} \) function, i.e., it is strictly increasing and $\gamma(0) = 0$. 
\end{definition}

A CBF $B$ specifies the controls that guarantee the {\em forward invariance} of $B_{\geq 0}$, i.e., all the trajectories of system
(\ref{eq:system}) that start in $B_{\geq 0}$ and follow controls that satisfy (\ref{eq:cbf-def-cond-ascent}), will remain 
within it 
at all times, as stated in the following theorem. If the set of unsafe states is disjoint from $B_{\geq 0}$ and the system starts from states in $B_{\geq 0}$, then the system can be kept safe by following such controls. 

\begin{theorem}[\cite{cbf_overview_2018}]
Any Lipschitz continuous control policy 
$\pi: \mathcal{D} \rightarrow \mathcal{U}$ where 
 $\forall x,  \pi(x) \in \{u \in \mathcal{U} : \nabla B(x)(f(x) + g(x)u) + \gamma(B(x)) \geq 0\}$
renders $B_{\geq 0}$ 
forward invariant. 

\end{theorem}

Given a reference controller \( \pi_{\text{ref}}: \mathcal{X} \rightarrow \mathcal{U} \) that does not necessarily guarantee safety, a CBF \( B \) can be used to filter its unsafe decisions. Specifically, a quadratic program (QP) can be formulated with the objective to find the closest safe control 
 to $\pi_{\text{safe}}(x)$, as follows (\cite{ames2016control}):

\begin{equation}\label{eq:qp}
    \begin{aligned}
    \pi_{\text{safe}}(x) &:= \arg\min_{u \in \mathcal{U}} \lVert u - \pi_{\text{ref}}(x) \rVert^2\quad 
        s.t. \quad \nabla B(x)(f(x) + g(x)u) + \gamma(B(x)) \geq 0. 
    \end{aligned}
\end{equation}

In our case, the state is a function of the non-interpretable representations of the image observations generated by the PVR backbones, as we will discuss later. The dynamics over such a state space are unknown, as it involves modeling uncertain, complex, and dynamic environments as well generating corresponding input images. This presents a challenge to traditional approaches for the synthesis of control certificates, particularly CBFs.  Fortunately, several works addressed the problem of learning barrier certificates for black-box dynamics  recently. As in \cite{ICRA_2025_pre-trained_vision_cbfs}, we use three of them, with few modifications when necessary, to construct our ensembles.
First, {\em In-Distribution Barrier Functions (iDBF)}
(\cite{castaneda2023distribution}) trains a control-affine dynamics model $\dot{x}=f_{\theta}(x)+g_{\theta}(x)u$ over the feature space of a variational auto-encoder using an offline dataset of safe trajectories. It also learns a behavior cloning  (BC) policy. The low probability actions under the distribution of that policy at the states visited in the dataset are assumed to result in unsafe states. These states along with the states visited by the safe trajectories are then used to train a CBF $B_\phi$. We use PVR backbones instead of training our variational auto-encoder. We also 
 have unsafe states in the dataset, those corresponding to  collisions, which we use instead of sampling ones from a BC policy. 
Second, {\em SABLAS}
 (\cite{qin2022SABLAS}) assumes that a simulator is available instead of an offline dataset. As iDBF,  
 it trains a nominal dynamics model, which it uses in training the CBF.  However, it accounts 
 for the discrepancy between the learned and the true dynamics in that process. We adapt the
  method to the setup where only offline trajectories, generated by the true dynamics, are available.  
 Finally, {\em Discriminating Hyperplanes (DH)}
 (\cite {lavanakul2024safety})
 directly trains a NN that maps states to hyperplanes  $a_\theta(x)^\top u = b_\theta(x)$ separating  safe and unsafe controls, generalizing 
 (\ref{eq:qp}).

\vspace{-0.3em}

\section{Method}\label{sec:methods}

In this section, we describe 
how we build our diverse 
ensembles 
and aggregate their outputs. 
\vspace{-0.3em}

\subsection{Designing member models of the ensembles}
\label{sec:Designing the member models of our ensembles}
We train each member vision-based safety filter using the approach described in \cite{ICRA_2025_pre-trained_vision_cbfs}. We use PVRs that resulted in the best performance in that study, which are: (1) CLIP (ViT-B/32 model) (\cite{clip}), a  model trained using contrastive learning on image-text pairs collected from the internet and (2) VC1 (ViT-B/32 model) (\cite{vc1}), a transformer-based encoder model pre-trained on data encompassing control and robotics tasks. In all of our experiments, we froze the backbones and only trained the safety filter heads and the layers aggregating the representations of the frames from the different cameras.
We consider the setting where a set of $M$ images $I^{1:M}_t$ are captured at each sampled time instant, 
e.g., by the various cameras mounted  on an autonomous vehicle. These images are fed one-at-a-time to a PVR backbone to obtain their respective representations $h_t^{1:M}$. We encode the identity of the camera capturing every image using positional encoding and concatenate it with its representation,  resulting in  
$
{h_t^{\prime}}^{1:M} := {h_t^i\ ||\ \mathrm{POS}(i)}_{i\in [1:M]}.
$
Then, we train an attention layer to compute 
$
score({h_t^{\prime}}^{i})
$,
which we use to create a unified representation using a weighted sum
$
h^*_t := \sum_{i=1}^M score({h_t^{\prime}}^{i}){h_t^{\prime}}^{i}.
$
We then define the system state as $x_t := \mathrm{MLP}_\theta(h^*_t,x_{t-1},u_{t-1})$, where $\mathrm{MLP}_\theta$ is a feedforward NN and 
$x_{t-1}$ and $u_{t-1}$ are the state and control at the previous time instant. 
Finally, we use iDBF, SABLAS, and DH methods to train the safety filters for the black-box dynamics over such a state space. 

\subsection{Aggregating the outputs of member models}\label{sec:model_aggregation}

This section discusses the methods we used to combine the outputs of the member models of the ensembles. We explored (weighted) averaging, majority voting, and consensus-based ensembles.

\paragraph{Weighted averaging-based ensembles} 
A simple approach to combine the outputs of the member models of an ensemble is to take their average. When they represent CBFs, the output of the ensemble can be defined as  $B_\text{ens}(x) := \sum_{i=1}^N w_i B_i(x)$, where $N$ is the number of member models, and $\forall i, w_i \geq 0$ and 
$\sum_{i = 1}^{N} w_i = 1$.  
One can either use uniform weights or  ones optimized using data. In either case, the left-hand-side of the constraint in (\ref{eq:qp}) becomes   
    $\dot{B}_\text{ens}(x, u) + \gamma (B_\text{ens}(x)) = \big(\sum_{i=1}^N w_i  \nabla B_i(x)(f_i(x) + g_i(x)u)\big) + \gamma\big(\sum_{i=1}^N w_i B_i(x)\big)$, where $f_i$ and $g_i$ represent the dynamics learned for  training $B_i$.  
    When the member models are DH-based ones, we can define a similar  constraint to that of the averaging-based CBF ensemble as follows: $\sum_{i=1}^N w_i (a_{i}(x)^\top u -b_{i}(x)) \geq 0$. Both constraints are affine in $u$, and one can still use  (\ref{eq:qp}) to obtain safe controls that follow the reference one. 
    As a separate note, as described in (\cite{lavanakul2024safety}), the output of a DH model represents a generalization of a CBF-based constraint. Particularly, $b_{i}(x)$ represents the term $-\nabla B_i(x)f(x) - \gamma(B_i(x))$ in the discriminating hyperplane defined by a CBF $B_i$ constraint. However, unless $\gamma$ is linear, the term 
    $-\sum_{i=1}^N w_i b_{i}(x)$ in the DH-based ensemble constraint  is different from the term $\big(\sum_{i=1}^N w_i  \nabla B_i(x)f(x)\big) + \gamma\big(\sum_{i=1}^N w_i B_i(x)\big)$ in the CBF-based one. For this reason, we only consider linear $\gamma$  in our experiments. That allows us to combine DH and CBF-based member models in the same ensemble. 

We optimize the weights $\{w_i\}_{i\in[N]}$, while freezing the member models.
We use this approach for both CBF and DH-based ensembles. 
We define the loss as:
    $\mathcal{L} = \mathcal{L}_\text{safe} + \lambda \mathcal{L}_\text{unsafe}$,
where 
$\mathcal{L}_\text{safe} = \sigma( - \dot{B}_\text{ens}(x, u) - \gamma (B_\text{ens}(x))) \cdot \mathbbm{1}(x' \in \mathcal{X}_\text{safe})$ or $\mathcal{L}_\text{safe} = \sigma( - \sum_{i=1}^N w_i (a_{i}(x)^\top u -b_{i}(x))) \cdot \mathbbm{1}(x' \in \mathcal{X}_\text{safe})$, $\mathcal{L}_\text{unsafe} = \sigma( \dot{B}_\text{ens}(x, u) + \gamma (B_\text{ens}(x))) \cdot \mathbbm{1}(x' \not\in \mathcal{X}_\text{safe})$ or $\mathcal{L}_\text{unsafe} = \sigma( \sum_{i=1}^N w_i (a_{i}(x)^\top u -b_{i}(x))) \cdot \mathbbm{1}(x' \not\in \mathcal{X}_\text{safe})$, $\sigma$ is the ReLU function, $x'$ is the  state appearing after $x$ in the trajectory, and $\lambda > 1$ is to further penalize miss-classifications of unsafe actions and handle dataset imbalance.  Nonetheless, for ensembles without DH-based models, one can alternatively choose to train  $\{w_i\}_{i\in[N]}$ to optimize both the values of the CBF on safe and unsafe states in addition to the hyperplanes classifying actions.

\paragraph{Majority voting-based ensembles}
The second approach we consider is to combine the outputs of the member models using 
majority voting. 
Each model decides whether a state or an action is safe or unsafe, and the final output is determined by the majority. 
To classify a state or action as safe, we should have strictly more votes for safety than unsafety, otherwise it is considered unsafe. For SABLAS and iDBF, we check if $ B_i(x) \geq 0$ to classify a state $x$ as safe and check if  $\dot{B}_i(x, u) + \gamma (B_i(x)) \geq 0$ to classify an action $u$ at state $x$ as safe. In the case of DH, a model does not classify states as it only defines hyperplanes separating actions to safe and unsafe ones. 
The constraint defined by the majority voting-based  ensemble
is not affine in $u$. Instead of the QP problem in (\ref{eq:qp}), the new optimization problem to find a safe action that is close to the reference can be formulated as  
a  Mixed-Integer Quadratic Program (MIQP), which is NP-complete (\cite{MIQP_is_NP_complete}). Solving such a problem is not suitable for real-time settings. Instead, if the majority voted that the {\em reference control} is unsafe, one can resort to a heuristic and select the models which voted in support of the decision and define a QP which have their constraints, and ignoring the other models.

\paragraph{Consensus-based ensembles}
In the final aggregation method that we use, ensembles have three member models: $M_1$, $M_2$, and $M_3$. We consider two cases.
In the first one, which we call the {\em specialized members}  case, we select $M_1$ and $M_2$ to be experts on different tasks: \( M_1 \) that is  
highly accurate in classifying safe actions and 
\( M_2 \) that is 
highly accurate in classifying unsafe ones. In the second case, which we call the {\em non-specialized members} one, 
\( M_1 \) and \( M_2 \) 
are both  equally capable in classifying both safe and unsafe actions. In both cases, we select \( M_3 \) to be 
an ensemble with higher accuracy in both classification objectives than $M_1$ and $M_2$. 
This aggregation method only calls \( M_3 \) when \( M_1 \) and \( M_2 \) disagree. The reason is that in the specialized members case, if $M_1$ decides that an action is safe and $M_2$ decides that it is unsafe, both decisions are in accordance with their expertise and we refer to $M_3$ to break the tie. Similarly, if $M_1$ decides that an action is unsafe and $M_2$ decides that it is safe, both cannot be trusted in their decisions as they are not experts, and again we refer to $M_3$. In the non-specialized members case, $M_3$ is breaking the tie among equally capable models.  

This method can be used to optimize computation time  
by 
calling the computationally 
expensive model only when needed. However, 
similar to the majority voting case, the constraint induced by the consensus-based aggregation method is not affine in $u$. One can follow a similar heuristic  
and check if $M_1$ and $M_2$ disagree on the safety of the reference control, then create a single constraint from $M_3$.
If they agree, then they can create a constraint from the model that is more accurate on the decision.

\section{Experimental Setup and Results }\label{sec:experiments}
We conducted several experiments to compare ensembles of 
safety filters 
with individual models on the  DeepAccident dataset (\cite{wang2024deepaccident}).  We show the results in Tables \ref{table:ensembles_and_individual_models} and \ref{table:EIR}. 

\subsection{Setup}

\paragraph{Dataset and data pre-processing}
DeepAccident (\cite{wang2024deepaccident}) is a synthetic dataset 
generated using CARLA simulating real traffic accidents 
reported by the National Highway Traffic Safety Administration (NHTSA), as well as safe driving scenarios. It includes action-annotated videos captured from six distinct cameras mounted on the ego vehicle with a total of 57k annotated frames. 
The control is a 2D vector determining the vehicle's velocity. We used the safety labels of the frames and the actions for the dataset which were created in \cite{ICRA_2025_pre-trained_vision_cbfs}. 
For each  trajectory with an accident, the first frame at which the collision happens was labeled as unsafe along with the frames following it. The five frames preceding the collision were labeled as safe, and the controls during that interval were labeled as unsafe. All other frames and controls were considered safe.

\paragraph{Evaluation metrics}
We use the classification accuracy of safe and unsafe states and actions. In addition, we use the {\em ensemble improvement rate (EIR)}, introduced in (\cite{theisen2024ensembles}), as a measure of improvement of the loss achieved by ensembles compared to  individual member models.
In the case of averaging-based ensembles, 
$\text{EIR} := \big(\frac{1}{N} \sum_{i\in [N]} L(f_i) - L(\bar{f})\big) / \big(\frac{1}{N} \sum_{i\in [N]} L(f_i)\big)$,
where $f(x)$ is $B_i(x)$ in the state classification task and $f(x,u)$ is $\nabla B_i(x)(f_i(x) + g_i(x)u) + \gamma\big(B_i(x)\big)$ in the action classification task. 
Also, $ \bar{f}(x)$ is 
$\sum_{i=1}^N w_i B_i(x)$
in the state classification task and $ \bar{f}(x,u)$ is 
$\big(\sum_{i=1}^N w_i  \nabla B_i(x)(f_i(x) + g_i(x)u)\big) + \gamma\big(\sum_{i=1}^N w_i B_i(x)\big)$
in the action classification task. The loss used in the EIR calculation for the action classification task is 
$\mathcal{L}(f) = \mathcal{L}_\text{safe}(f) + \lambda \mathcal{L}_\text{unsafe}(f)$, defined in Section \ref{sec:methods} with $\lambda=18$. For the state classification task, it is modified so that $\mathcal{L}_\text{safe}(f) := \frac{1}{|D|} \sum_{x \in D}\sigma(- f(x)) \cdot \mathbbm{1}(x \in \mathcal{X}_\text{safe})$
 and
$
\mathcal{L}_\text{unsafe}(f) := \frac{1}{|D|} \sum_{x \in D} \sigma(f(x)) \cdot \mathbbm{1}(x \not\in \mathcal{X}_\text{safe})
$, where $D$ is the set of states in the test set  and  $\mathcal{X}_\text{safe}$ is the set of safe states. We replace the averages over $D$ with the averages over all pairs $(x,u)$ in the test set and $\mathcal{X}_{\text{safe}}$ to the set of safe state-action pairs when considering the action classification task. 
In the case of majority voting-based ensembles, we have the same definition of EIR, but consider $\bar{f}$ to be the majority voting ensemble and $L(f)$ is the zero-one loss, i.e., is zero when a state or an action is   
correctly classified.

\begin{table}[htbp]
\scriptsize
\setlength{\tabcolsep}{3pt} 
\centering
\begin{tabular}{l|l|l|c c c|c c c}

\toprule
Aggr. method & Training Method & Backbone & Safe States & Unsafe States & EIR & Safe Actions & Unsafe Actions & EIR\\
\midrule

\multirow{15}{*}{\makecell{\hspace{4mm}Majority\\\hspace{4mm}voting}}&
\multirow{3}{*}{iDBF} 
& CLIP & 90.27 & 64.35 &4.73& 90.98 & 48.91&0.92 \\
                      
&& VC1 & 75.74 & 80.43 &4.50 & 77.06 & 73.10&6.20 \\

&& VC1-CLIP & 82.99 &82.17&20.25& 84.152& 69.29 &18.53\\
\cmidrule{2-9}

& \multirow{3}{*}{SABLAS} 
& CLIP & 84.94 &69.57 &3.79& 85.16 & 59.51&9.63 \\
        
&& VC1 & 74.38 & 78.26 &5.93& 75.23 & 68.75&4.49 \\

&& VC1-CLIP & 80.94& 81.74& 20.11&82.23& 69.84 &20.15\\
\cmidrule{2-9}

& \multirow{3}{*}{DH}
& CLIP & - & -&- & 30.19 & 94.29&9.59 \\
                      
&& VC1 & - & - & - &23.86 & 98.37 &6.50\\
&& VC1-CLIP & - & -&- & 19.02 & 99.46&2.58\\
\cmidrule{2-9}
& \multirow{3}{*}{SABLAS-iDBF-DH}
 & CLIP & - & -&- & 86.44 & 62.50 &23.79\\
&& VC1 & - & - & -&70.19 & 81.25&24.85\\
& & VC1-CLIP & - & -&- & 76.99 & 80.98&36.77\\
\cmidrule{2-9}

& SABLAS-iDBF & VC1 & 73.28 & 81.30&3.31 & 75.38 & 73.37 &8.57\\
\midrule

\multirow{15}{*}{\makecell{\hspace{3mm}Averaging\\\hspace{3mm}$(w_i=\frac{1}{N})$}}
& \multirow{3}{*}{iDBF} 
& CLIP & 91.80  & 63.04 &12.00 & 92.47& 48.10 &6.96 \\
                      
&& VC1  & 76.13 & 77.39&17.36  & 77.32 & 69.84 &16.97 \\

&& VC1-CLIP  & 89.99 & 68.26 &40.68  & 90.68  & 53.2 & 30.35\\
\cmidrule{2-9}

& \multirow{3}{*}{SABLAS} 
& CLIP & 90.58  &61.74 &30.98 & 88.94  & 51.90 &19.74\\
        
&& VC1  & 74.65  & 77.83 &26.44 & 76.39  & 68.75&25.47 \\

&& VC1-CLIP  & 88.57& 63.91&46.47& 89.29& 54.08&33.55\\
\cmidrule{2-9}

& \multirow{3}{*}{DH}
& CLIP & - & - &-& 30.24 & 86.41&29.06  \\
                           
&& VC1  & - & -&- & 20.37  & 99.18&12.5  \\

&& VC1-CLIP  & - &-& -& 24.65& 97.28&33.89\\
\cmidrule{2-9}

& \multirow{3}{*}{SABLAS-iDBF-DH}
  & CLIP & - & - &-& 91.00 & 51.09&23.13\\
& & VC1 & - & - &-& 78.48 & 69.02&37.3 \\
& & VC1-CLIP & - & - &-& 90.47 & 53.53 &38.89\\
\cmidrule{2-9}

& SABLAS-iDBF & VC1 & 75.48 & 76.52 &23.42& 78.53 & 67.93&35.81 \\
\midrule

\multirow{5}{*}{\makecell{\hspace{3.5mm}Weighted\\\hspace{3.5mm}averaging}} 

&iDBF &VC1& 74.18& 79.57&14.35&   75.23  & 73.37&18.96  \\
\cmidrule{2-9}
&SABLAS &VC1& 72.61& 79.57&20.06&   74.50  & 72.01 &24.22\\
\cmidrule{2-9}
&SABLAS-iDBF-DH &VC1& -&  - &-&  78.12  & 68.48 &30.45 \\
\cmidrule{2-9}
&SABLAS-iDBF &VC1& 76.56&  77.83 & 24.47 &79.08  &72.01&39.48  \\
\midrule

\multirow{3}{*}{\makecell{\hspace{3mm}Consensus\\\hspace{3mm}based}}
&Specialized members &VC1-CLIP   &-&- &-  & 76.37$\pm$2.53 & 76.88$\pm$3.64   &- \\

\cmidrule{2-9}
&\makecell{Non-specialized \\ members}   &VC1-CLIP &-&-  &-  & 77.59$\pm$2.70          & 75.75$\pm$3.54 &-\\

\midrule

\multirow{7}{*}{\makecell{ \hspace{4.5mm}Member\\\hspace{4.5mm}models}} 
& \multirow{2}{*}{iDBF} 
& CLIP & 88.98$\pm$4.2 & 64.09$\pm$4.23 & -&89.65$\pm$4.12 & 49.56$\pm$6.9 &- \\
&& VC1 &74.55$\pm$7.10 & 79.56$\pm$5.91 &-& 75.72$\pm$7.07 & 71.14$\pm$6.63  &-\\
\cmidrule{2-9}

& \multirow{2}{*}{SABLAS} 
& CLIP & 85.66$\pm$4.7 & 54.95$\pm$5.66 & -&82.84$\pm$5.87 & 55.86$\pm$9.04  &\\
&& VC1 & 73.46$\pm$7.64 & 75.56$\pm$7.45 &-& 74.29$\pm$5.61 & 67.06$\pm$6.55  &\\
\cmidrule{2-9}

& \multirow{2}{*}{DH}
& CLIP & - & -&- & 35.43$\pm$7.04 & 82.17$\pm$8.12  &-\\
&& VC1 & - & - &-& 28.16$\pm$15.95 & 89.51$\pm$6.58  &-\\
\midrule

\multirow{3}{*}{\makecell{Large\\single models\\(increased width)}} 
&iDBF & VC1 & 94.17$\pm$1.34 & 52.03$\pm$8.76 &-& 94.89$\pm$1.28 & 38.77$\pm$8.35 &- \\
\addlinespace[3pt]
\cmidrule{2-9}
&SABLAS & VC1 & 91.90$\pm$1.24 & 57.68 $\pm$ 6.35 & -& 91.95$\pm$1.46  & 46.55$\pm$5.89 &\\

\addlinespace[3pt]
\midrule
\multirow{3}{*}{\makecell{Large\\single models\\(increased depth)}} 
&iDBF & VC1 & 88.37$\pm$4.83 & 60.96$\pm$18.27 &-& 88.81$\pm$4.72 & 50.54$\pm$16.13 &- \\

\cmidrule{2-9}
&SABLAS & VC1 & 80.70$\pm$11.86 & 64.13$\pm$12.99 & -& 82.25$\pm$12.09  & 57.45$\pm$14.82 &\\

\addlinespace[3pt]

\bottomrule
\end{tabular}

\caption{Performance of ensembles with different aggregation methods, PVR backbones, and training methods compared to the performances of their member models and large non-ensemble models. }\label{table:ensembles_and_individual_models}

\end{table}

\subsection{Results and analysis}

We trained five models with different weight initializations and hyper-parameters for every pair of  a backbone and a safety filter training
method to create all of our ensembles. When designing an ensemble that has a certain pair, we include all of the five corresponding trained models.
For example, 
the ensembles using three training methods and one backbone
are composed of fifteen individual models.

Table \ref{table:ensembles_and_individual_models} shows  the accuracies and EIR of single-backbone and multi-backbone ensembles using various output aggregation and safety filter training methods. It also includes the accuracies of the member models and individual large models with comparable total parameters to the ensembles. 
When reporting the results of individual models, we use the average accuracies taken over five models along with their standard deviation. Hereafter, when presenting action classification accuracy percentages, we use the format (safe action classification accuracy $\%$, unsafe action classification accuracy $\%$), unless stated otherwise. We focus more on the action classification task in our analysis as it is the fundamental purpose of the safety filter.

\paragraph{Comparison of ensembles and individual models}
By comparing the results of ensembles and member models in  Table~\ref{table:ensembles_and_individual_models}, we observe that the former generally perform equally or better than the average of the latter, as expected. For example, the weighted averaging-based and majority voting-based ensembles of models with a VC1 backbone trained using  
iDBF achieve action classification accuracies of (75.23, 73.37) and 
(77.06, 73.10), respectively, which are slightly better than the average of their members (75.72, 71.14). Similarly, member models trained using SABLAS 
and having a
VC1 
backbone have an average action classification accuracy of
(74.29, 67.06) while their corresponding 
weighted averaging-based ensemble 
has a better accuracy of 
(74.50, 72.01). 

In the cases with the CLIP backbone, both member models and the ensembles demonstrated relatively low performance. For example, SABLAS with CLIP member models achieve an average accuracy of (82.84, 55.86), which increases to (85.16, 59.51) for the majority voting-based ensemble.
This shows that while ensembles help balance or improve performance,
if the underlying individual models perform poorly, the ensemble’s performance is also likely to be limited. 

 When we look at 
 ensembles with more diverse member models, such as the majority voting-based one using 
 all training methods (iDBF, SABLAS, DH) and 
 both VC1 and CLIP backbones, we observe an accuracy of   (76.99, 80.90), which is better than the average results of all of its member models.

Most weighted averaging-based ensembles, utilizing variations of the SABLAS and iDBF training methods along with a VC1 backbone, consistently achieve accuracies in the range of 74–80\% on classifying safe actions. This marks an improvement compared to the range of 74–76\% accuracy achieved by the average of  member models. Similarly, these ensembles demonstrate 72–74\% accuracy on classifying unsafe actions, outperforming the member models, which achieve 67–71\%.

Even the least effective ensembles, using the averaging aggregation method with uniform weights, show slight benefits by improving both safe and unsafe state/action classification accuracies compared to member models, or by enhancing one metric while causing only a minor decrease in the other.

\paragraph{Comparison of single- and multiple-backbone ensembles}

Multiple-backbone ensembles outperform single-backbone ones, which can be attributed to the diversity of features captured.
VC1 is trained with masked autoencoding on egocentric video datasets from diverse robotic simulators and tasks (including navigation) as well as ImageNet. 
On the other hand, CLIP is trained with contrastive learning on image-text pairs, captures features that complement those from VC1.

We can observe in Table~\ref{table:ensembles_and_individual_models} that the majority voting-based ensemble trained using SABLAS achieves action classification accuracies of (75.23, 68.75) with the VC1 backbone, and (85.16, 59.51) with the CLIP one. 
When it has both the VC1 and CLIP backbones, i.e., some of its members use VC1 and the others use CLIP, it achieves  (82.23, 69.84), capturing approximately the best performance on each metric from the single-backbone ensembles.
Moreover, the majority voting-based ensemble using iDBF achieves state classification accuracies of (90.27, 64.35) with CLIP and (75.74, 80.43) with VC1 while
with both VC1 and CLIP, 
it achieves (82.99, 82.17), improving the unsafe states classification accuracy and achieving the average safe states classification  one over those of  the single-backbone ensembles. 
The same trend of improving or preserving performance metrics  can be seen for all training methods on both states and actions classification accuracies. 
Finally, the EIR for ensembles using both VC1 and CLIP is always larger than those using only CLIP or only VC1. The only exception to this trend is the majority voting-based ensembles using DH.

\paragraph{Comparison of different consensus-based 
ensembles}\label{sec:mixed_learners}
For specialized members, where single models M1 and M2 are highly accurate in either safe or unsafe predictions, we used one of the five filters we trained with SABLAS and CLIP as a safe action classification expert 
and one of the five filters we trained with DH and CLIP as an unsafe action classification one. Those trained with iDBF and CLIP and those trained with DH and VC1 are viable alternatives for these tasks.
For non-specialized members, we used one of the five filters we trained with iDBF and VC1
and one of the five filters trained with SABLAS and VC1. M1 and M2 would be two of the models in these sets.
M3 is the majority voting-based ensemble using   
VC1 and CLIP, all 
training methods (SABLAS, iDBF, DH), and having an accuracy of (76.99, 80.90).
We used all combinations of M1 and M2 models (50 experiments: 25 for the specialized case and 25 for the non-specialized one).

While reducing the computational demands of using a large 
ensemble $M3$ all the time, the consensus-based aggregation method decreases the classification accuracies compared to $M3$.
Employing an ensemble with balanced \( M_1 \) and \( M_2 \) models (non-specialized members) achieves comparable performance to specialized \( M_1 \) and \( M_2 \) models while requiring significantly fewer calls to \( M_3 \).

The specialized members ensemble achieved an average accuracy of (76.37, 76.88), but required frequent calls to \( M_3 \) (58.92\% of the time). In contrast, the non-specialized members ensemble called \( M_3 \) only 21.67\% of the time, achieving a similar accuracy of (77.59, 75.75). 

\paragraph{Comparison of model aggregation  methods}
The choice of aggregation methods 
has a substantial impact on the ensembles performance. Both majority voting and weighted averaging-based ensembles significantly improve performance.
Weighted averaging provides more consistent results.
 Corresponding ensembles trained using SABLAS and iDBF with VC1 consistently achieve between 74\% and 80\% safe action accuracy and between 72\% and 74\% unsafe action accuracy. However, the best ensemble was trained using majority voting and achieved (76.99, 80.98). 
Uniform averaging-based ensembles were the least effective, as they reflect the average performance of their individual models rather than leveraging the unique strengths of each member.  Weighted averaging partially solves this issue by weighing better performing models higher, but the weights are determined during training and frozen during deployment, making the weights input-independent.
Consensus-based method combine the expertise of both models by inferring from their disagreement on a given input a potential classification error that requires another expert opinion. In our case, the expert is a majority voting-based ensemble by itself. Finally, majority voting suppresses anomalies in an input-dependent manner, i.e., the models that it ignores change for every input, and results in the best performance. Majority voting is also easier to implement and does not require the extra step of learning the weights.

\paragraph{Comparison of large models and ensembles}
We compare the performance of ensembles with larger individual models.
Our aim is to investigate if the improved accuracy of the ensembles is caused by their large number of parameters or from other characteristics, such as the diversity of  member models.  
We trained two versions of large models, both having VC1 as a backbone,  but one using SABLAS and the other using iDBF. 
We increase the size of the safety filters by (1) increasing the number of hidden layers and (2) increasing the number of neurons per hidden layer.

We trained ten models with five hidden layers, versus the  two hidden layers in member models, and ninety five neurons per layer, versus the sixty four in member models. These deeper models had approximately eight times the number of parameters as the original ensemble members. 
We also trained wider models with two hidden layers and  220 neurons per layer. These models had roughly ten times the parameters of the ensemble members but performed poorly in comparison to the deeper models, as shown in Table~\ref{table:ensembles_and_individual_models}.
Thus, we focus our analysis on the deeper models.
The larger models' performances in  classifying unsafe actions remain suboptimal compared to ensembles. 
The large models trained using SABLAS achieve an improvement in safe action classification accuracy (82.25\%)
over similar member models (74.29\%) and corresponding ensembles (74.5-76.4\%), but fail to balance this with unsafe action accuracy, reaching only 
57.45\%, while the similar member models achieve 67.06\% and corresponding ensembles achieve 68.75-72\%. 
This trend is evident across both deeper and wider large models, for both SABLAS and iDBF.

Notably, the smaller models outperform the larger ones. This might be because the larger models tend to overfit the limited size of the training dataset. 
To the best of our knowledge, DeepAccident is one of the largest datasets currently available in the literature with diverse accident scenarios.  %
\paragraph{Comparison of ensembles on in-distribution (IND) and out-of-distribution (OOD) data}
We considered four towns from the DeepAccident dataset as IND data and withheld the data from the  remaining three towns as OOD data.
We trained fourteen  models on the training dataset portion of the IND data and created their uniform  averaging and majority voting-based ensembles and  computed their EIR, as shown in Table \ref{table:EIR}.
Each town has a different environment, but they share similar accident patterns and trajectories. Thus, such a configuration provides  a limited view on the the performance of the ensembles on OOD data.
EIR remains consistently positive for OOD test data, though slightly lower than for IND test data. This indicates that ensembles maintain advantage over member models on OOD samples, though slightly reduced 
compared to IND data.

\begin{table}[h]
\scriptsize
\centering
\begin{tabular}{l c | c c | c c}
\toprule
\multicolumn{2}{c}{} & \multicolumn{2}{c}{Majority voting} & \multicolumn{2}{c}{Averaging} \\
\midrule
Training method & Test Set & EIR States & EIR Actions & EIR States & EIR Actions \\
\midrule
\multirow{2}{*}{SABLAS}
& IND & 19.06\% & 10.7\% & 37.6\% & 23.1\% \\
& OOD & 9.68\% & 9.37\% & 35.00\% & 22.97\% \\
\midrule
\multirow{2}{*}{iDBF}
& IND & 11.5\% & 2.4\% & 18.8\% & 12.9\% \\
& OOD & 5.2\% & 0.87\% & 15.2\% & 8.3\% \\
\bottomrule
\end{tabular}
\caption{IND and OOD EIR for ensembles with a VC1 backbone.}\label{table:EIR}
\end{table}

\vspace{-2em}
\section{Conclusion}
We conducted an extensive analysis of various ensemble configurations, including multiple perception backbones  (VC1 and CLIP), different safety filter training methods (SABLAS, iDBF and DH), diverse weight initializations and hyper parameters as well as various model aggregation methods (averaging, weighted averaging, majority voting and consensus-based).
Our results showed that ensemble methods consistently improved performance and out-of-distribution generalization of safety filters compared to both member models of the ensemble and to larger single models with comparable number of parameters as the ensemble.


\bibliography{references}

\appendix

\end{document}